\pdfoutput=1

\documentclass[11pt]{article}

\usepackage{EMNLP2022}

\usepackage{times}
\usepackage{latexsym}

\usepackage[T1]{fontenc}

\usepackage[utf8]{inputenc}

\usepackage{microtype}

\usepackage{inconsolata}

\usepackage{multirow}
\usepackage{soul, color}
\sethlcolor{green}
\usepackage{paralist}
\usepackage{graphicx}

\newtheorem{example}{Example}
\title{Bootstrapping NLP tools across low-resourced African languages: an overview and prospects}

\author{C. Maria Keet \\
  Department of Computer Science \\
  University of Cape Town \\
  South Africa \\
  \texttt{mkeet@cs.uct.ac.za} \\}

\begin{document}
\maketitle
\begin{abstract}
Computing and Internet access are substantially growing markets in Southern Africa, which brings with it increasing demands for local content and tools in indigenous African languages. Since most of those languages are low-resourced, efforts have gone into the notion of bootstrapping tools for one African language from another. This paper provides an overview of these efforts for Niger-Congo B (`Bantu') languages. Bootstrapping grammars for geographically distant languages has been shown to still have positive outcomes for morphology and rules or grammar-based natural language generation. Bootstrapping with data-driven approaches to NLP tasks is difficult to use meaningfully regardless geographic proximity, which is largely due to lexical diversity due to both orthography and vocabulary. Cladistic approaches in comparative linguistics may inform bootstrapping strategies and similarity measures might serve as proxy for bootstrapping potential as well, with both fertile ground for further research. 
\end{abstract}

\section{Introduction}

Nearly 1.5 billion people live in Africa, of which many who speak multiple languages other than the relatively well-resourced languages English, French, and Arabic, among an estimated 1441 to 2169 African languages \citep{Hammarstrom18}. Notably, by first language speakers, Swahili of the Niger-Congo family is the next-largest language (around 50 million L1 speakers, some 100-200 million overall), with as close second the Afroasiatic Hausa and then Oromo, and then Yoruba and Igbo in the Niger-Congo family with each around 28 million L1 speakers\footnote{Numbers from various open and paywalled sources, collated at \url{https://en.wikipedia.org/wiki/Languages_of_Africa}.}. While illiteracy exists in Sub-Saharan Africa, there are still very many people who can read and write in indigenous languages, some of which  having official status in one or more countries where they are used in education, work, and social life. This entails a need for language support for task such as spelling and grammar checking, translation, and natural language generation. Within the context of the United Nations'  Sustainable Development Goals\footnote{\url{https://sdgs.un.org/goals}}, they include language technologies to ameliorate the language gap (and consequent lower service \citep{Hussey13}) in healthcare \cite{Byamugisha17,Marais20}, educational digital assistants to ease the burden of overworked teachers in crowded classrooms of up to 100 learners \citep{Keet21ist}, and many other supportive tasks for society, such as machine translation for humanitarian response \citep{Oktem21}. In this paper, we zoom in on on the Niger-Congo B (`Bantu'), or NCB, family of languages. In \citet{Joshi20}'s classification of NLP support for languages, the NCB languages fall into three of the six categories: the ``left behinds'', the ``scraping-bys'', and the ``hopefuls'', with isiZulu (spoken in South Africa) and Swahili (spoken primarily in Tanzania and Kenya) in the latter group.

A selection of the usual NLP tasks have been taken up for a  few languages of the Niger-Congo family, notably indeed Swahili and isiZulu, and to a lesser extent Yoruba, Igbo, isiXhosa, and Runyankore. Examples are diverse. They range from corpus creation for data-driven NLP, such as the IsiZulu National Corpus \citep{Khumalo15} that was used for a statistical language model for a spellchecker \citep{Ndaba16}, the Mashakane grassroots initiative\footnote{\url{https://www.masakhane.io/}} that focuses on data-driven machine translation for multiple African languages \citep{Nekoto20},  to data-driven text-to-speech \cite{Marais20} based on Qfency\footnote{\url{http://www.qfrency.com/}}, and other language modelling and data augmentation (e.g., \citep{Byamugisha20,Mesham21}; see also \citet{Kambarami21} for an overview). The main knowledge-driven approaches include terminology development in general \cite{Khumalo17} and domain-specific (e.g., \cite{Engelbrecht10}), and rule-based morphological analysers \cite{Pretorius03,Bosch17}, grammars \cite{Bamutura20,Pretorius17}, and natural language generation \cite{KK16lre,Byamugisha19,MK20siz}. Most of the research has taken place over the past 5-10 years and is gaining pace, albeit still for only a slowly increasing number of NCB languages.

The low-resourced and very low-resourced languages\footnote{While there is no crisp demarcation of `low' in low-resourced languages, it is to be understood as having only small (e.g., 20K tokens) or no curated monolingual or parallel corpora, limited (including outdated) or no grammar books, and typically also comparatively few researchers and funding.} face a `catch-22', however: there are few language resources but one needs language and linguistics resources to increase the language resources. A well-known idea is to try to `bootstrap' resources for a very low-resourced language from a low-resourced one; e.g., to bootstrap a spellchecker for isiNdebele from an isiZulu spellchecker, which are both languages in the Nguni group  of the NCB languages in South Africa.

Theoretically this makes sense, but practically it is nontrivial to figure out bootstrapping potential and strategies. In this paper, we report on a preliminary review of published research on bootstrapping for the NCB languages to provide better insight into it such that it can better inform NLP tasks for NCB languages. Observing that it remains imprecise as to what should be assessed quantitatively to gauge bootstrapping potential, the research demonstrates that bootstrapping with rules and grammars extends to more languages than initially assumed and with data-driven approaches to fewer languages due to limited lexical proximity due to variations in vocabulary and orthography.

In the remainder of the paper, we first summarise NLP-relevant features of NCB languages and linguistics-focussed categorisations of NCB languages in Section~\ref{sec:langintro}. We then proceed to the key questions for bootstrapping and the review in Section~\ref{sec:boot}, discuss in Section~\ref{sec:disc}, and close in Section~\ref{sec:concl}.

\section{NCB languages: some key features}
\label{sec:langintro}

This section describes language and linguistic of NCB languages insofar as they are relevant to computation thus far, with first key language features and then the grouping of subsets of NCB languages.

\subsection{Grammar and orthography}

The system of noun classes is emblematic for the NCB languages. Each noun belongs to a noun class and there are up to 23 noun classes; see Table~\ref{tab:ncs} for a summarised overview. All NCB languages retained the lower numbers and are fairly similar up to noun class 11, in that the odd numbered classes contain nouns in the singular and even number classes nouns in the plural, they pair up mostly in the same way, and they have a large overlap in the kind of things that can be found in each pair of noun classes. After that, the NCB languages diverge on which noun classes are retained in the language and there may not be a singular/plural pairing. For instance, isiZulu's {\em ubuntu} `humanity' is an abstract concept in noun class 14 for which there is no singular or plural, whereas in Chichewa a noun in noun class 14 may have a plural in noun class 6.
The nouns with their noun classes govern a rich system of concordial agreement across sentence constituents, ranging from verb conjugation to modifying adjectives, possessives, and other relational notions.

\begin{table*}[h]
\centering
\caption{Generalisation of the semantics of the kinds of objects that the nouns in the respective noun classes (NCs) refer to. Examples from isiZulu (1-11, 14, 15), Chichewa (12,13,16-18), Hunde (19), Runyankore (20,21), and Luganda (22,23).  
(Source: adapted from \cite{BKD18}.)}
\begin{tabular}{p{0.7cm}|p{6.2cm}|p{5.5cm}} \hline
\textbf{NCs} & \textbf{Semantics (generalised)} & \textbf{Examples} \\ \hline
1 &\multirow{2}{*}{People and kinship}  & {\em umfana} (nc1) `boy' \\ 
2 &  & {\em abafana} (nc2) `boys' \\ \hline
3 & \multirow{2}{*}{\parbox{6.2cm}{Plants, nature, some parts of the body}}  & {\em umuthi} (nc3) `tree' \\ 
4 &  & {\em imithi} (nc4) `trees' \\ \hline
5 & \multirow{2}{*}{\parbox{6.2cm}{Fruits, liquids, parts of the body, loan words, paired things}} & {\em ikhala} `nose' \\ 
6 &  & {\em amakhala} `noses'  \\ \hline
7 & \multirow{2}{*}{Inanimate objects}  & {\em isihlalo} `chair' \\ 
8 & & {\em izihlalo} `chairs' \\ \hline
9 & \multirow{2}{*}{Loan words, tools, and animals} & {\em inja} `dog'\\ 
10 & & {\em izinja} `dogs' \\ \hline
11 & \multirow{2}{*}{\parbox{6.2cm}{Long thin stringy objects, languages, inanimate objects}} & {\em ucingo} `wire'\\ 
(10) & & {\em izingcingo} `wires' \\ \hline
12 &  \multirow{2}{*}{Diminutives}  & {\em kagalimoto} `small car' \\ 
13 &  &{\em timagalimoto} `small cars' \\ \hline
14 & Abstract concepts & {\em ubuhle} `beauty' \\ \hline
15 & Infinitive nouns & {\em ukucula} `to sing' \\ \hline
16 & \multirow{3}{*}{Locative classes} & {\em pamsika} `round the market' \\ 
17 &  &  {\em kumsika} `at  the market' \\ 
18 &  & {\em mumsika} `in the market' \\ \hline
19 & Diminutives & {\em hy\`und\`u} `a little bit of porridge' \\ \hline
20 & \multirow{3}{*}{Augmentative and pejorative} & {\em ogusajja} `big ugly man'\\ 
21 &  & {\em agasajja} `big ugly men'\\ 
22 & & {\em gubwa} `mutt' (pejorative of dog)\\ \hline
23 & Locative class & {\em eka} `at home'\\ \hline
\end{tabular}
\label{tab:ncs}
\end{table*}

The verbs have a so-called ``slot system'' where each slot fulfils a specific function, if used \cite{Khumalo07}: there are eight ordered slots, being the pre-initial, initial, post-initial, pre-radical, (verb) radical, pre-final, final, and post-final slot. The pre-initial and post-initial can take tense, aspect, mood and negation, and the pre-final can take tense, aspect, mood and valence change (causative, accusative, reciprocative, and passive). The initial is for the subject concord to conjugate the verb depending on the subject in the sentence and the pre-radical slot is for the object concord. The final slot is for the final vowel (e.g., default /a/ in isiZulu, but /i/ if the verb is negated) and post-final is used for extensions including the wh-questions and locative suffix.

Many of the NCB language are agglutinating and thus have a substantial set of phonological conditioning rules especially for vowel coalescence and vowel elision. For instance, for `(located) in the envelope' in isiZulu, one has to modify {\em imvilophu} `envelope' with phonologically conditioned locative prefix ({\em e-}) and suffix ({\em -ini}) to result in {\em emvilophini}, and the enumerative `and' {\em na-} merges with the successive noun, as in, e.g., (na- + umfana =) {\em nomfana} `and the boy' and (na- + inja =) {\em nenja} `and the dog'.

They overwhelmingly use Latin script, with some also Arabic script and modern indigenous writing systems. Among the ones that use Latin script, there can be language-specific variations; e.g.,  isiZulu typically does not have words with an /r/ and Swahili no /q/, Mboshi has an $\varepsilon$ variant of /e/ in addition to the /e/, and there are letter combinations to stand in for more consonants, such as a `hard b' and a `soft b' (/bh/ and /b/, respectively), and for `click sounds', such as a nasalised click that may written as /nc/. Many NCB languages are tonal, although this may not be reflected in the orthography \cite{Maddieson19}.

\subsection{Categorising NCB languages}
\label{sec:comp}

\begin{figure}[t]
\centering
\includegraphics[width=0.49\textwidth]{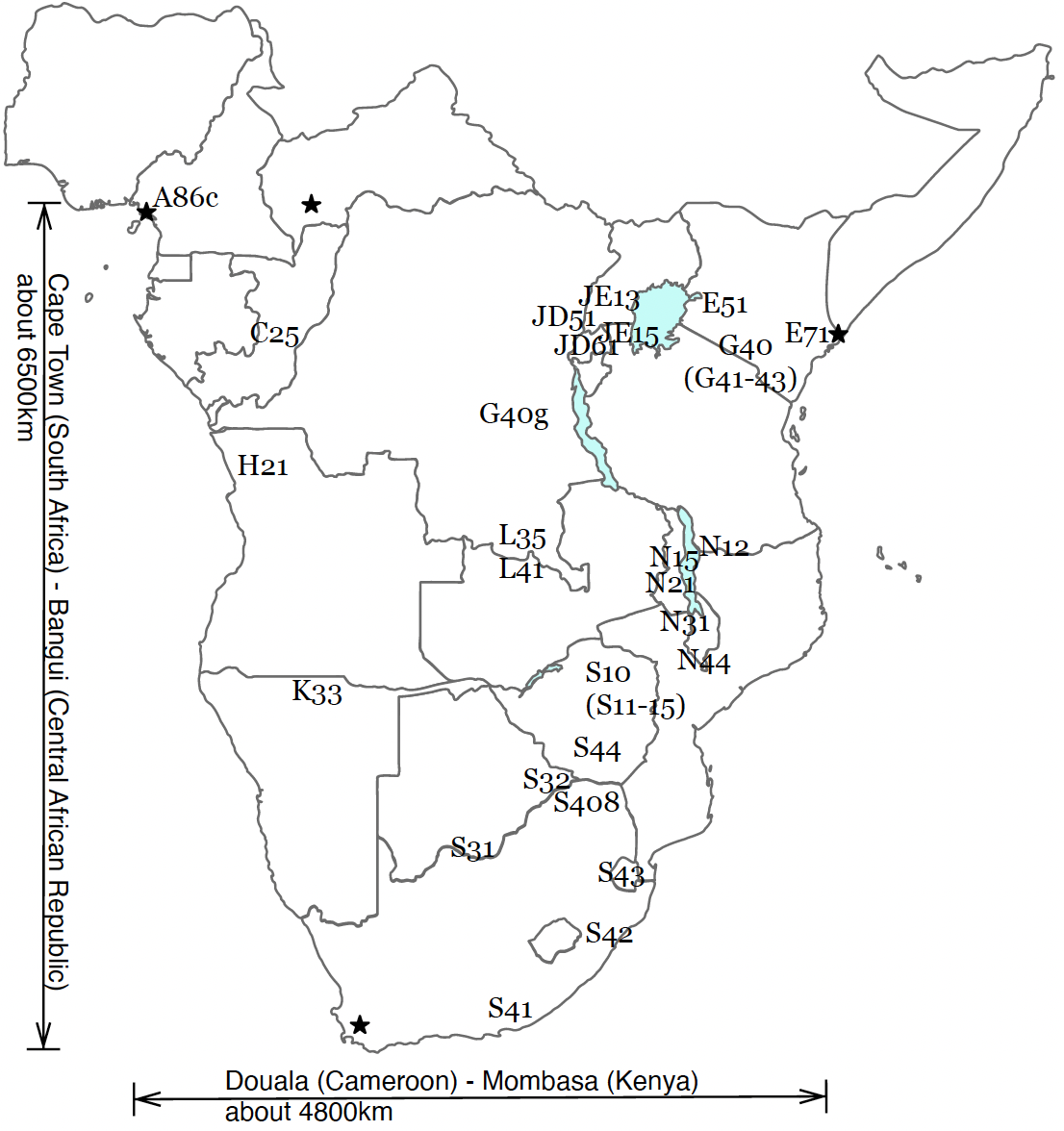}
\caption{Map of Southern Africa where NCB languages are spoken. The NCB languages mentioned in this paper are included on the map with their respective Guthrie zone code (geographic extent not indicated); see Table~\ref{tab:exLangList} for further detail.}
\label{fig:map}
\end{figure}

There have been multiple attempts at grouping the NCB languages according to various parameters. The one most well-known is based on geographic regions devised by \citet{Guthrie71}, which has been updated in \cite{Moho03} and again informally in 2009 with detailed maps and many references\footnote{The document is available at \url{https://brill.com/fileasset/downloads_products/35125_Bantu-New-updated-Guthrie-List.pdf} (last accessed 3 Sep 2022), but does not seem to have been published.\label{fn:moho}}. The system counts from A to S, from top-left in Cameroon to down-right in South Africa. Each zone has groups, indicated by increments of 10 (e.g., A10), 
where all languages within the group have arbitrarily ordered increments of 1 (e.g., A11 and A12), and possibly further minor increments, such as A111 and A12a. A map overlaid with the NCB languages mentioned in this paper is shown in Fig.~\ref{fig:map} (the non-NCB Niger-Congo languages Yoruba and Igbo are located left of A86c, in Nigeria); 
see also Table~\ref{tab:exLangList}.

In addition to descriptive linguistics-based overviews, comparisons, and high-level groupings \citep{Guldemann18}, some of those research efforts also have a computational component that also may inform potential for bootstrapping  language resources. \citet{Petzell13} compared 8 languages in Tanzania on 27 morphophonological and morphosyntactic parameters, which showed that the neighbour-joining trees for lexical and for grammatical distance are  distinct to the extent that they indicate different language clusters, and, consequently, a merger of the two can lead to different clusters. This feature was also observed in the cladistic analyses of 87 languages by \citet{Rexova06}. They  merged 92 lexical and 52 grammatical characters and tried two different classification computations (parsimony with Bremer indices and Bayes), which revealed different trees and putative percentages for similarity. Notably, it also showed that, while the clades largely cluster by the high-level Guthrie zones, this does not hold in all cases (e.g., Pokomo (E71) and Swahili (G40 (G41-43)) are siblings in both trees), and multiple times not so for groups (e.g., Sanga (L35) and Kaonde (L41) are siblings). Similar works exist, varying input and algorithms; e.g., also sound files and an algorithm based on DNA sequence optimisation to avoid orthographic peculiarities in computational comparisons \citep{Whiteley19}. Such works could provide guidance for devising a bootstrapping strategy for a particular NLP task.

\section{Bootstrapping: questions and related work}
\label{sec:boot}

The notion of bootstrapping resources from one language for another applies to any set of languages. General questions that direct the research that surface include, mainly:
\begin{enumerate}
\item Bootstrapping for which NLP task? 
	\begin{enumerate}
	\item Does the type of task (e.g., machine translation (MT) versus natural language generation (NLG)) and approach (corpus-based, rules-based, class of algorithm) matter?
	\item If so: how and to what extent? If no: why not?
	\end{enumerate}
\item How to measure language similarity, as a proxy for bootstrappability potential? 
	\begin{enumerate}
	\item Which metrics to use or devise? 
	\item For a given metric, what is the threshold (or fuzzy function) for `similar' and for determining `similar enough for bootstrapping anyhow' and for `similar enough for bootstrapping for NLP task X'? 
	\item In which way do the similarity metric and bootstrapping ease relate, if at all?
	\item Can metrics for distinguishing (separating) languages automatically be flipped and be used for similarity for bootstrapping, and if so, how? 	
	\end{enumerate}
\item Are bootstrapping gains reliably quantifiable beyond a `many hours saved' or `any amount of a better BLEU score' in carrying out NLP task X? 
\item To what extent, if at all, does some success of cross-language learning (e.g., using a L1 corpus for training an L2 spell checker) shed any useful light on bootstrappability prospects of L2 from L1?
\item If bootstrapping from L1 to L2 works well, would it work just as well from L2 to L1? Or: what is the `fastest' route from L$_i$ to L$_j$?
\end{enumerate}

These broad questions are hard to answer based on the available research on NCB languages, but progress is being made. 

\begin{table*}[!h]
\centering
\caption{Listing of the languages mentioned in the paper, with Guthrie zone and ISO 639-3 codes for at least some of the languages grouped within it, the NLP task(s) where the language was mentioned (if applicable) and whether it was used in some way of bootstrapping (``Boot.'') efforts. MT: machine translation; NLG: natural language generation; POS: part-of-speech; IR: information retrieval.}
\begin{tabular}{p{3.1cm}|p{2.3cm}|p{2.2cm}|p{5.5cm}|p{0.7cm}} \hline
\textbf{Language} & \textbf{Guthrie code} & \textbf{ISO 639-3} & \textbf{Task mentioned} & \textbf{Boot.}  \\ \hline
Chichewa   & N31 & nya & IR, similarity & -- \\
 Cinyanja    & N31a? & nya &  IR, similarity & -- \\
  Cisena   & N44 & seh & IR, similarity & -- \\
  Citonga  & N15& tog & IR, similarity &  -- \\
  Citumbuka   & N21& tum & IR, similarity & -- \\      
G\~{\i}k{\~u}y{\~u} & E51 & kik & prefix extraction & + \\
Hunde   & JD51 & hke & -- & --  \\
Kaonde & L41 & kqn & (computational cladistics) & -- \\
 Kimbundu  & H21 & kmb & similarity & -- \\
Kinyarwanda & JD61 & kin &noun class prediction & + \\
Kwangali  & K33 & kwn & (computational cladistics) & --  \\
 Luganda     & JE15 & lug & noun class prediction & +  \\
isiNdebele (ZW)  & S44 & nde &  morphological analysers & +  \\
isiNdebele (ZA)   & S408 & nbl & pronunciation dictionary & +  \\
isiXhosa  & S41 &xho & NLG, MT, morphological analysers, pronunciation dictionary, similarity & + \\
 isiZulu  & S42 &zul & NLG, morphological analysers, MT, prefix extraction, POS tagger, corpus development, spellchecker,  pronunciation dictionary, similarity & + \\
  Mboshi & C25 &mdw & -- & -- \\
Mpiemo  & A86c &mcx & POS tagger & + \\
Ngoni  & N12 &ngo & morphological analysers & +  \\
Pokomo  & E71 & pkb & (computational cladistics) & --  \\
 Runyankore  & JE13  & nyn & NLG, similarity, (computational cladistics), noun class prediction & +  \\
Sanga & L35  &sng & (computational cladistics) &  --  \\
 Sepedi  & S32 &nso & POS tagger, pronunciation dictionary, similarity & + \\
Setswana   & S31 &tsn & morphological analysers, pronunciation dictionary, similarity & + \\
 Shona    & S10 (S11-15) &sna (twl, mxc, twx, ndc) & MT,  similarity & +  \\
siSwati    & S43 & ssw & morphological analysers & +  \\ 
 Swahili   & G40 (G41-43)  & swa, swh (ccl, sta) & MT, POS tagging, pronunciation dictionary, news item monitoring, similarity, (computational cladistics) & + \\ 
 Swahili (Congolese)  & G40g & swc & MT & + \\
 \hline
\end{tabular}
\label{tab:exLangList}
\end{table*}

The literature review was necessarily of an exploratory nature rather that that of, e.g., a meta review, since only limited research on bootstrapping for NCB languages has been carried out (discussed in Section~\ref{sec:disc}). To cast the net wide, we searched the scientific literature by 
\begin{compactitem}[-]
\item multiple terminology variants: including bootstrapping, bootstrappability, bootstrapping NLP tools, cross-language;
\item broad language terms, such as African languages, Bantu languages, Niger-Congo languages;
\item specific language names with variant spellings, such as both Swahili and Kiswahili, Zulu and isiZulu; and
\item a number of categories of NLP tasks, including machine translation, morphological analyser.
\end{compactitem}
They were combined in multiple combinations of these dimensions, such as ``bootstrapping morphological analyser Swahili''. Of the papers found, the search was then augmented with `forward' searches in time to find other papers that cited them, which were then assessed manually on whether those papers were also about bootstrapping for NCB languages. 

The literature we found describe few and largely disparate efforts, which we nudged to attendant topics that may assist with bootstrapping across NCB languages: bootstrapping within a single language and similarity among NCB languages. We grouped them into two approaches: qualitative with estimates on time saved and quantitative with similarity measurements and output assessments. We describe and discuss them in the next two subsections, and subsequently synthesize and discuss them in Section~\ref{sec:disc}.

\subsection{Qualitative assessments}

Qualitative assessments can be grouped into bootstrapping across language families, cross-language rules- or grammar-based approaches among the NCB family, and within-language bootstrapping. We'll discuss each in turn.

\subsubsection{Experiments across language families} Reusing tools for resource-rich languages for bootstrapping tools for low-resourced languages in another language family is  a tempting idea---that one's NLP tool would be that much generalisable. Among others, \citet{Baumann14} tried to find isiZulu prefixes based on an English wordlist under the assumption of loanword adoption and \citet{Pauw07} tried prefix extraction for G\~{\i}k{\~u}y{\~u} with the Indo-European languages-tailored AutoMorphology. The latter compared AutoMorphology to a new maximum entropy modeling (maxent) algorithm, which was much better suited to the agglutinating languages than AutoMorphology. Cross-family efforts have obtained only  limited results; rather, they fed into the drive to develop NLP tools for the NCB languages proper and the notion of learning from other NCB languages that are relatively more closely related. For instance, common spellcheckers for languages such as English are dictionary-based, but it is infeasible to use that approach for highly inflectional agglutinating languages---or: it was easily bested with character 3-grams extracted from a small corpus \citep{Ndaba16}. For the Europe Media Monitor's inclusion of Swahili, the bootstrapping amounted to only determining which NLP sub-tasks needed to be carried out, but tools had to be sourced, adapted, and developed as well (in about 3 person-months) \citep{Steinberger11}, as was the case for word alignment in an English-Swahili parallel corpus \cite{Pauw11}. Once \citet{Pauw11} had created that word alignment, however, POS tagging for Swahili was bootstrapped from the POS tagged aligned English with 74.8\% accuracy.

\subsubsection{Rules and grammar bootstrapping} The qualitative experiments on bootstrapping across NCB languages focus on the knowledge-based approach. This was first reported for morphological analysers \citep{Bosch08,Pretorius09} on similarity among languages in the Nguni language cluster vs disjunctive orthography of the Tswane cluster in South Africa, where the within-cluster was substantially faster for bootstrapping. For isiZulu, it was estimated to cost around 3000 highly
skilled person hours, but subsequent analyser development for the three other Nguni languages (isiXhosa, siSwati and isiNdebele) based on ZulMorph took only about 300 hours in total \citep{Bosch08}. 
At the sentence level, bootstrapping natural language generation was experimented with for rules for Runyankore (JE13, spoken in Uganda) from rules for isiZulu (S42, South Africa, about 4800km further south). This was successful, albeit indicating only ``a lot easier'' \cite{Byamugisha16}. Their later work further demonstrated that Guthrie zones indeed are not a reliable predictor for bootstrapping advantages \cite{Byamugisha19}. 

A linguistic analysis across Guthrie zones, in this case for bootstrapping morphological analysers for ``Zimbabwe Ndebele'' (S44, isiNdebele) and Ngoni (N12, Tanzania) from the isiZulu (S42) Zulmorph tool also concluded that bootstrapping across zone is feasible, albeit easier for the one closer-by (isiNdebele) than geographically further away (Ngoni), due tot influences of more disparate NCB languages on morphemes and noun classes \citep{Bosch12}, as is the case for Runyankore cf. isiZulu as well. But, like with \citet{Byamugisha16}'s outcomes, also for these language comparisons, it appeared that word categories and the order of morphemes agree largely \citep{Bosch12}. 

In all these experiments, precise measurements beyond rough estimates are inherently difficult to obtain, because the development task for the language to bootstrap from has too many variables to make it reliable to devise one final  precise value, be this for how to compute grammar similarity or the time saved. The time saved also depends on linguistic knowledge of the developers and quality and recency of the grammar documentation, and the bootstrap values are sensitive to how one translates qualitative input into a numerical value or vector.
We illustrate this in the next example. 

\begin{example}
{\em Consider ontology or knowledge graph verbalisation of the all-some axiom pattern ($C \sqsubseteq \exists R.D$): the template `pattern' for isiZulu and Runyankore are, respectively: 
\begin{compactitem}
\item[Zu:]   [QCall] [$C$] [SC][$R$] [$D$] [RC][QC]dwa 
\item[Nyn:] [QCall] [$C$] [SC][$R$] hakiri [$D$] [RC]mwe
\end{compactitem}
as generalised from the example and algorithm of \citet{Byamugisha16}, with [QCall] the quantitative concord for the universal quantifier, [SC] the subject concord, [RC] the relative concord, and [QC] the quantitative concord for the existential quantifier. QCall, SC, RC, and QC are clitics governed by the noun class of the respective noun.

The structure of the sentences and the components are mostly the same, with just variation in vocabulary. The only difference is in part of the verbalisation of ``$\exists$'': a fixed {\em hakiri} versus a noun class-dependent [QC], whilst [RC], {\em dwa}  and {\em mwe} are the same elements. A sample calculation for syntax could be a percentage for each element in common (100\%) and the ordering similarity computed with a Levenshtein distance for elements or some other measure (5 of the 8 are in the same position, or 62.5\%). It would thus score very low on lexical proximity, possibly high depending on which metric is chosen for the grammar, and a merged value base don equal weighting could ever be only at most 50\%. That was still enough for bootstrapping advantages. \hfill $\diamondsuit$}
\end{example}

A high grammar similarity clearly can assist bootstrapping rules-based NLP approaches and tasks, whereas the substantive lexical distance due to entirely different vocabulary will negatively affect any cross-lingual data-driven NLP task.

\subsubsection{Bootstrapping within a single language} 

Although this sense of bootstrapping is not the scope of the paper, it is worth mentioning that it has been done in the comparatively early years of language technologies for South African languages, upon which some of the later techniques build at least in part. An illustrative example concerns data-driven text-to-speech applications with minimal human-in-the-loop intervention, where first a pronunciation dictionary for German was attempted \citep{Davel03} that was then extended to experiments with South African languages  \citep{Davel09} that evolved into afore-mentioned Qfrency that, in turn, was embedded in the Awezamed app \citep{Marais20}.
They commenced with little data and expanded it through a continuous feedback loop rather than relying on similarities of grapheme-to-phoneme of various subsets of the 11 official South African languages and rather than resource-consuming corpus development. 
The term `bootstrapping' is used likewise elsewhere for, among others, Sepedi, isiZulu, Swahili, and Mpiemo NLP resources \citep{Prinsloo05,Spiegler10,Hammarstrom08,Schlippe14}. 
Each commenced with limited data and then used rules-based and/or statistical approaches with a human-in-the-loop for curation in the iterations.  Neither reported on relative gains after each step in the procedure nor did they examine the effect of the input type or size (e.g., varying the amount of seed words or rules) on the quality of the output. Hence,   determining a sweet spot for an ideal, or at least minimum, amount of seed data to obtain a certain level of quality remains an open question.

\subsection{Quantitative measurements}

With the increasing popularity of data-driven approaches, measurements have gained attention, but quantitative measurements are scarce and even then with different foci and mostly on similarity rather than assessments on the bootstrapping process. We'll discuss both.

\subsubsection{Data-based bootstrapping}

A `small data' data-based approach informed by grammar was taken by \citep{Byamugisha22} in the task of predicting the noun class of nouns for Runyankore, Luganda, and Kinyarwanda. Sub-word, word-level, and syntax level with nearest neighbour strategies were used, which showed a 10\% improvement in noun class prediction in the case of cross-language resource reuse from a Runyankore resource to Luganda. This was attributed to noun class prefix similarities: 15 Luganda noun class prefixes out of 21 are the same as for Runyankore. For Kinyarwanda, which has only 5 exact matches in prefixes with the Runyankore ones, combining the resource had a detrimental effect of about 8\% \cite{Byamugisha22}. 

Concerning `large data' data-based approaches, various machine and deep learning-based MT and similar tasks are investigated for and across NCB languages with bootstrapping and transfer learning. They are largely based on informal assumptions, ranging from, informally, `it works for Swahili so your language should work too, since both are African languages'---i.e., linguistically uninformed---to the usual ones among a handful of languages. 
Regarding the latter, for instance, \citet{Nyoni21} compared English-to-isiXhosa, to isiZulu and to Shona with different neural models and then tested English-to-isiZulu with BLEU. Better BLEU scores were obtained in transfer learning from isiXhosa to isiZulu (a gain of 6.1$\pm$0.4 from the baseline of 8.7$\pm$0.3) rather than from Shona to isiZulu (a gain of 0.9$\pm$0.8), and likewise they observed an expected improvement combining the small parallel datasets of isiZulu and isiXhosa on English-to-isiZulu performance. Different amounts of BLEU score improvements indirectly indicate a similarity estimate, especially regarding lexical proximity, and offer quantitative values for this type of bootstrapping. Those values may, however, be skewed by the size of the training data if they are not roughly equal in size for the languages tested. 

\begin{table*}[t]
\centering
\caption{Illustration of differences in orthography and similarity of constituents (adapted and extended from \citet{Kambarami21}) with the agglutinating Shona and isiZulu and disjunctive Setswana and English, and Shona and isiZulu morphology (right) if it were to have been written disjunctively.}
\begin{tabular}{p{2.3cm}|p{1.7cm}|p{2cm}|p{2.3cm}||p{2.5cm}|p{2.3cm}} \hline
\textbf{Shona} & \textbf{isiZulu} & \textbf{Setswana} & \textbf{English} & \textbf{Shona morphology} & \textbf{isiZulu morphology}  \\ \hline
 ndinomutya & ngimsaba  & kea mo tsaba & I fear him/her & ndino mu tya & ngi m saba \\ 
 ndinomuda & ngimthanda & kea mo rata & I love him/her &  ndino mu da & ngi m thanda\\ 
 ndinomuziva & ngimbona & kea mo tseba & I know him/her &  ndino mu ziva & ngi m bona\\ 
ndinomubatsira & ngimsiza & kea mo thusa & I help him/her & ndino mu batsira & ngi m siza \\ \hline
\end{tabular}
\label{tab:exOrtho}
\end{table*}

For MT, scraping texts in any African language together and using a try-and-see approach to what BLEU outputs for the task is popular, and any training data in any of the languages---even from different language families---shows some improvement for any model. \citet{Ogundepo22}'s results showed that English-Swahili and Swahili-English MT had the highest BLEU scores among the Niger-Congo languages, but for the other languages, it does not provide any bootstrapping potential insight. This because of training data size differences (the number of tokens and sentences trained on was 10 to 100-fold more for Swahili than most of the other texts of the other languages), inclusion of languages in different language families (e.g., also Hausa, Somali), and the BLEU scores were often still in the single digit range. Most of these MT efforts are to/from English, occasionally French \citep{Oktem21} (to Congolese Swahili, G40g). MT among NCB languages is investigated even less\footnote{We did not find any research, but this may be due to the search terms, which zoomed in on bootstrapping.}, even though that also will be useful, such as for providing assistance to an isiZulu-speaking doctor explaining home care to a Setswana-speaking patient---just like there are direct translators in other languages families, like a Spanish to Italian direct translation rather than devising a Spanish-to-English-to-Italian detour.

\subsubsection{Measuring similarity}
 
A different strategy is to measure similarity among NCB languages somehow and to extrapolate those values to an informed guess for bootstrappability. 
 
\citet{Keet16tr} assessed orthographic similarity with various measures for 10 African languages in different Guthrie zones, including Swahili, Kimbundu-Mbundu, Runyankore, Shona, and several  South African languages. A key insight with just the cumulative frequency distributions of the words in the Universal Declaration of Human Rights was that not all NCB languages are agglutinating and three groups can be identified. There are those that indeed are highly agglutinative, such as  Runyankore, Shona, and isiZulu, those that are as disjunctive as English, such as Swahili, and those that are highly disjunctive, such as Sepedi---and statistically significantly so. Based on the morphological analysis reported by \citet{Pretorius09} and the comparison table by \citet{Kambarami21}, the latter two clusters may still have a similar enough grammar for bootstrapping compared to starting from scratch, but they mainly will have fewer phonological conditioning rules due to the disjunctive orthography. This is illustrated and elaborated on in the following example. 

\begin{example}\label{ex:orth}
{\em Consider the examples in Table~\ref{tab:exOrtho}, where agglutinating Shona and isiZulu have 4 types and 4 tokens in the examples, and disjunctive Setswana and English have 12 tokens and 6 types. Let's assume a similarity calculation with the Levenshtein distance: from {\em ndinomutya} to {\em ngimsaba} it is already 6 and from {\em kea mo tsaba} to {\em ngimsaba} it is 7; it is as bad or downhill from there with the other examples up to a distance of 10. An agglutinating {\em keamotsaba} to {\em ngimsaba} distance (or, vice versa, the disjunctive {\em ngi m saba}) reduces it  to 5 and {\em keamothusa} to {\em ngimsiza} from 10 to 8, which are still not good, but it demonstrates that orthographic conventions matter: a 20-30\% improvement in similarity.\footnote{In contrast to the large lexical distances observed, the respective grammar components are the same: SC-OC-VerbStem, i.e., first the subject concord (verb conjugation) for `I', then object concord for `him/her' (the OC does not encode gender of the human), and then the verb stem.}  \hfill $\diamondsuit$}
\end{example}

Conversely, a quick check on amount of agglutination may provide a rough idea of good bootstrapping potential, as supported also by the Runyankore, isiNdebele, and Ngoni resources bootstrapping from isiZulu \cite{Byamugisha16,Bosch12}. Whether this simple proxy holds generally is yet to be investigated. 

The phonological conditioning does have an effect on the ruleset for understanding and generating text. This was examined in the context of similarity measures for verbs in isiZulu and isiXhosa \cite{MK19}. For the domain of verbs for weather forecasts, the 52 isXhosa and 49 isiZulu rules have 42 rules in common, but it reached only 59.5\% overall similarity on an adapted Driver-Kroeber metric as morphosyntactic similarity measure, although 99.5\% for the past tense only \citep{MK19}. In other words: it fluctuates substantially to either not be a reliable indicator or one where the notion of `sufficient similarity' has to have a low threshold value. 

Around the same time, \citet{Chavula17} tried a weighted metric that uses the Ordered Weighted Aggregator, on geographically neighbouring languages Chichewa (N31) and Citumbuka  (N21), which outperformed Dice on morpheme segmentation. This was then extended for five languages in Malawi (Cinyanja, Citumbuka, Chichewa, Citonga and Cisena) and zooming in on predicting intelligibility to serve reranking of results for better  information retrieval \citep{Chavula21}. Those intelligibility prediction values using Random Forest classifiers achieved the best results with a prediction accuracy up to nearly 0.75 for the largest data set size, for selected topics. 
Intelligibility entails similarity of vocabulary and grammar and therewith bootstrapping for any NLP task and cross-lingual learning in data-driven NLP. 
Also here, the meaning of the actual value of prediction accuracy is not fully clear. That is, it leaves questions regarding the grounding, in the sense of whether that 0.75 is good enough for understanding text sufficiently or with difficulty or ease, and whether a few percentage points more or less would have any effect on, e.g., statistical language modelling and resulting performance of, say, data-driven spellcheckers or machine translation.

\section{Discussion}
\label{sec:disc}

Revisiting the typical sort of bootstrapping questions from the beginning of Section~\ref{sec:boot} in the light of the related work, one has to concede that very few questions can be answered conclusively. There are many different NLP tasks and a wide variety of these tasks have been attempted for one or more NCB languages over the years with both data-driven and rules-based approaches or a combination thereof. And for each of those tasks, limited bootstrapping has been experimented with to a limited extent. 
We tried to be inclusive in the review of papers on NCB languages, but since there are very many NCB languages known under even more names that are not always similar strings\footnote{e.g., Hima and Runyankore or Nyanya and Chichewa, rather than only the common prefix addition or skipping like with isiZulu and Zulu or ciShona and Shona.} (and many NLP tasks) that are sometimes only mentioned in a paper's section on data collection, this overview on bootstrapping for NCB languages is likely to be incomplete. Nonetheless, several insights can be distilled from it already.

\begin{figure}[t]
\centering
\includegraphics[width=0.49\textwidth]{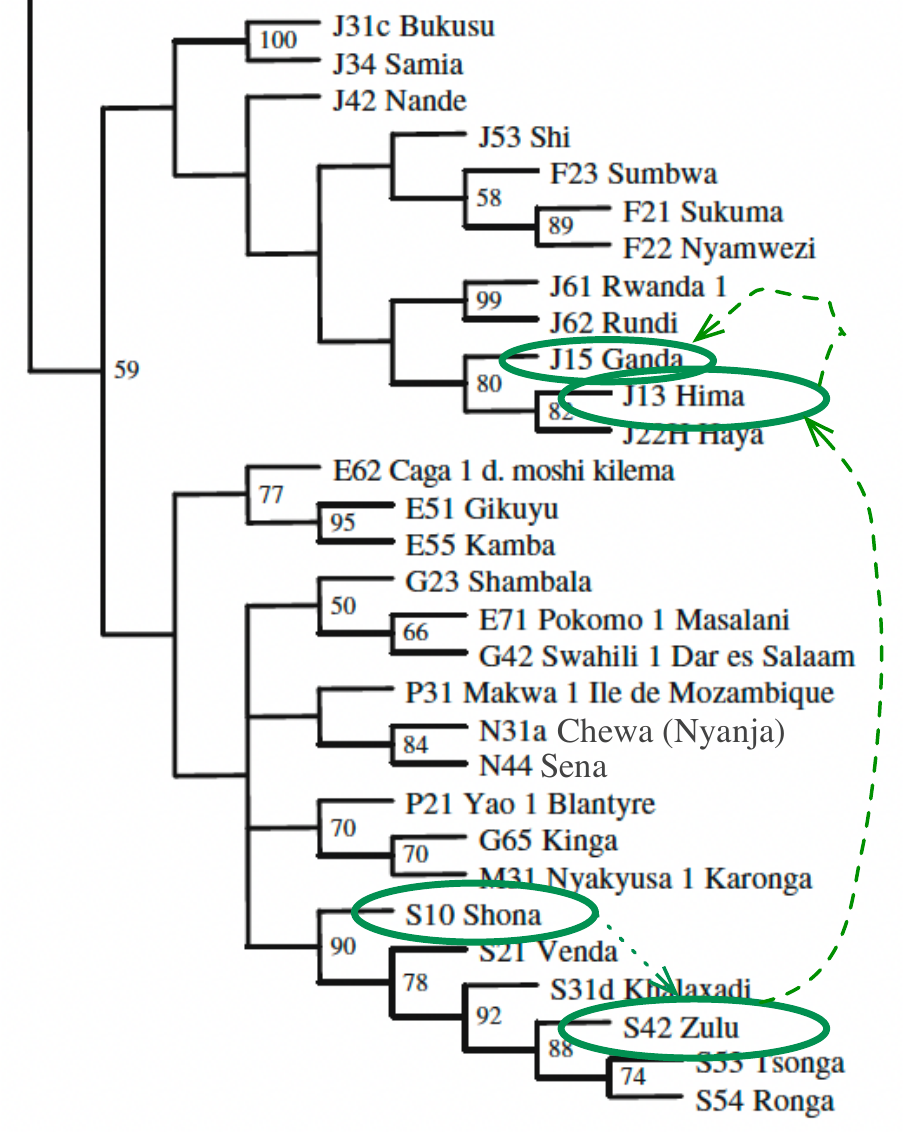}
\caption{Section of \citet{Rexova06}'s cladogram-like binary tree with bootstrapping attempts encircled for the overlap in language inclusion: success of grammar from S42 (isiZulu) to J13 (Runyankore (Hima)), mixed syntax and data from J13 to J15 (Luganda), and limited benefits of enriching S42 data with S10 (Shona) data.}
\label{fig:clade}
\end{figure}

The rules-based bootstrapping experiments focussing on grammar and morphology demonstrated that many of the key grammar components and their use are similar across NCB languages in different Guthrie zones thousands of kilometres apart, as demonstrated for isiZulu on the one hand and Runyankore and Ngoni on the other. Their distance is visualised in the cladogram-like diagram of \citet{Rexova06} in Fig.~\ref{fig:clade}, suggesting that at least the languages in this sub-branch are promising targets for bootstrapping for grammar-based tasks and approaches. Conversely, the negligible advantage of cross-language learning of adding Shona data for English-isiZulu MT, yet being relatively nearby, is not at all promising. Data augmentation with an arbitrary other African language is even less likely to boost BLEU scores. An illustration that suggests why, was seen in the small example in Example~\ref{ex:orth} and more careful selection of a neighbouring language likely will yield better results. 
Other issues (see, e.g., \cite{Hovy16}) are yet to be dealt with as well, such as which data is used for bootstrapping---bible translations and old texts are not representative for current language use, as shown by \cite{Ndaba16}.

Clade-based approaches to grouping NCB languages, such as in Fig~\ref{fig:clade} and mentioned in Section~\ref{sec:comp}, may further guide the `guessing game' to better informed experimenting of bootstrapping. For instance, whether it is indeed harder, and if so how much so, to bootstrap an NLP task for, say, Kwangali (K33), which is located in a separate branch from the one shown in Fig~\ref{fig:clade}, from isiZulu as compared to all the current reported isiZulu-to-LanguageX bootstrapping results.

Last, language similarity computed with the various measures as a proxy for bootstrappability did not reveal more than that a low similarity value on lexical proximity may still yield good results for rules-based bootstrapping. This is invariant of the similarity measures that have been used. Better grounding of similarity measures to the quality of executing of a specific NLP task may give more insight into the meaning of the values obtained. The notions of similarity for bootstrapping may avail of the concept of intelligibility for which many methods exist \cite{Gooskens13}. While of interest, it was deemed beyond the scope, because it requires the existence of substantial resources.

\section{Closing remarks}
\label{sec:concl}

The review of bootstrapping efforts for Niger-Congo B (`Bantu') languages showed that a variety of bootstrapping efforts have been investigated to a very limited extent. The most promising for bootstrapping are rules/grammars, even if the languages spoken are geographically distant. Lexical diversity due to orthographic and vocabulary differences indicated that geographic proximity cannot be relied on for enhancing data-driven approaches to NLP tasks. The results on measuring similarity as proxy for estimating bootstrapping leaves ample room for future research, both on the metrics and the grounding thereof for reliable estimation. Also, more cross-fertilisation between linguistic theory and comparative linguistics and computational experiments on bootstrapping may have an added value for guiding what to bootstrap from what.

\section*{Acknowledgements}
This work is based on the research supported in part by the National Research Foundation
of South Africa (Grant Number 120852).

\bibliography{keet,keetrefs}
\bibliographystyle{acl_natbib}

\end{document}